\documentclass[runningheads]{llncs}
\usepackage{graphicx}

\usepackage{tikz}
\usepackage{comment}
\usepackage{color}
\usepackage{url}  
\usepackage{graphicx}
\usepackage{amsmath,amssymb,amsfonts}
\usepackage{booktabs}
\usepackage{bbm}
\usepackage{tabularx}
\newcolumntype{L}[1]{>{\raggedright\arraybackslash}m{#1}}
\newcolumntype{C}[1]{>{\centering\arraybackslash}p{#1}}

\usepackage[accsupp]{axessibility}  

\begin{document}
\pagestyle{headings}
\mainmatter
\def\ECCVSubNumber{4921}  

\title{Hierarchical Semi-Supervised Contrastive Learning for Contamination-Resistant \\ Anomaly Detection} 


\titlerunning{HSCL: Contamination-Resistant Anomaly Detection}
%
\author{Gaoang Wang\inst{1} \and
Yibing Zhan\inst{2} \and
Xinchao Wang\inst{3} \and
Mingli Song\inst{1} \and
Klara Nahrstedt\inst{4}}
\authorrunning{G. Wang et al.}
%
\institute{Zhejiang University, China \and
JD Explore Academy, China \and
National University of Singapore, Singapore \and
University of Illinois at Urbana-Champaign, USA\\
\email{gaoangwang@intl.zju.edu.cn, zhanyibing@jd.com, xinchao@nus.edu.sg, brooksong@zju.edu.cn, klara@illinois.edu}}
\maketitle

\begin{abstract}
Anomaly detection aims at identifying deviant samples from the normal data distribution. Contrastive learning has provided a successful way to sample representation that enables effective discrimination on anomalies.
However, when contaminated with unlabeled abnormal samples in training set under semi-supervised settings, current contrastive-based methods generally 1) ignore the comprehensive relation between training data, leading to suboptimal performance, and 2) require fine-tuning, resulting in low efficiency. 
To address the above two issues, in this paper, we propose a novel hierarchical semi-supervised contrastive learning~(HSCL) framework, for contamination-resistant anomaly detection. 
Specifically, HSCL hierarchically regulates three complementary relations: sample-to-sample, sample-to-prototype, and normal-to-abnormal relations, enlarging the discrimination between normal and abnormal samples with a comprehensive exploration of the contaminated data. Besides, HSCL is an end-to-end learning approach that can efficiently learn discriminative representations without fine-tuning. 
HSCL achieves state-of-the-art performance in multiple scenarios, such as one-class classification and cross-dataset detection. Extensive ablation studies further verify the effectiveness of each considered relation. The code is available at \url{https://github.com/GaoangW/HSCL}.
\end{abstract}

\section{Introduction}

Anomaly detection aims to distinguish outliers from in-distribution samples.
In addition to the basic image classification task that aims at identifying abnormal visual samples from the base class \cite{tack2020csi,han2021elsa}, anomaly detection is also widely exploited in other fields, such as defect detection \cite{akcay2018ganomaly,kim2020gan,chen2021unsupervised,li2021cutpaste} and abnormal event detection \cite{liu2018future,gong2019memorizing,wang2019anomaly,lv2021learning}. Some works focus on designing anomaly scores and anomaly classifiers, such as \cite{pang2021explainable,qiu2022latent}. Some methods combine reinforcement learning \cite{pang2021toward} and knowledge distillation \cite{ma2022deep} in the anomaly detection.
Some self-supervised anomaly detection methods \cite{tack2020csi,kim2020gan,salehi2020arae} use clean normal data in training and achieve much progress. Recently, more and more works \cite{gornitz2013toward,kingma2014semi,ruff2019deep,han2021elsa} focus on the contaminated setting, where unlabeled abnormal samples are included in the training set, following the semi-supervised framework. This setting is much closer to the real situations that training data may be contaminated by abnormal samples, while a small labeled set can be easily acquired. 

\begin{figure}[!t]
\begin{center}
\includegraphics[width=0.9\linewidth]{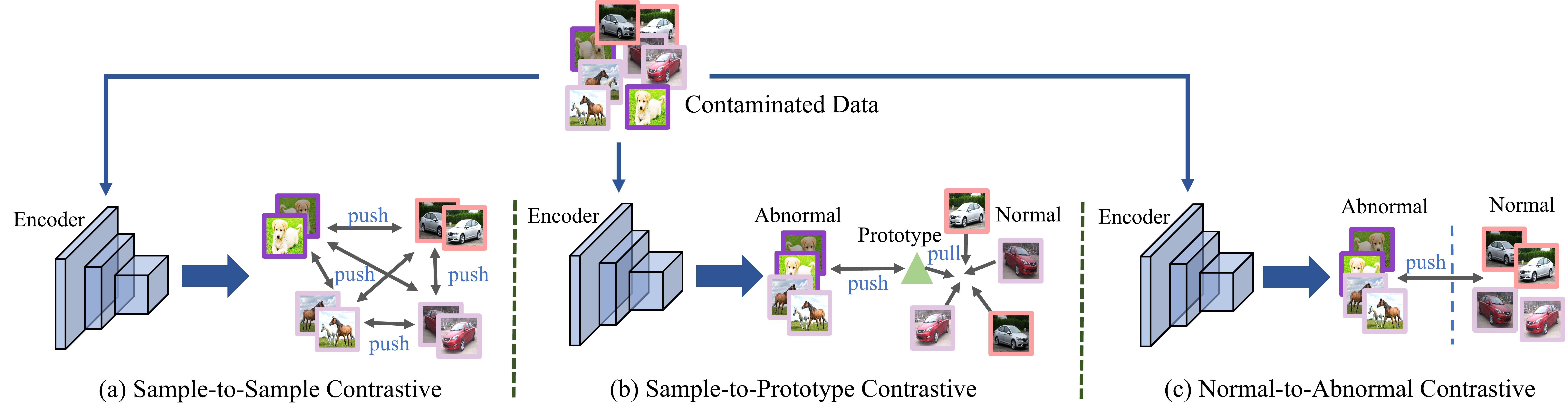}
   \caption{Hierarchical contrastive relations with contaminated data 
   for anomaly detection. Light red, purple, and light pink colors represent labeled normal, abnormal, and unlabeled data, respectively. The green triangle represents the class prototype of normal samples. Current contrastive learning-based anomaly detection approaches only consider (a)~sample-to-sample relation and largely overlook the discrimination between normal and abnormal samples. Our proposed method jointly accounts for (a)~sample-to-sample, (b)~sample-to-prototype, and (c)~normal-to-abnormal contrastive relations with both labeled and unlabeled data.}
\label{fig:representation}
\end{center}
\end{figure}

With limited labeled information under the semi-supervised setting, a good representation learning strategy is always crucial to identify the abnormal samples. Inspired by the recent success of contrastive learning for visual representation \cite{chen2020simple,he2020momentum,caron2020unsupervised,grill2020bootstrap,li2021contrastive,zhong2021graph}, much progress of contrastive learning has been made in anomaly detection, and contrastive learning-based approaches \cite{tack2020csi,sohn2020learning,reiss2021mean,han2021elsa} have significantly outperformed the conventional reconstruction-based anomaly detection approaches \cite{zhou2017anomaly,zhao2017spatio,gong2019memorizing,liu2018future,kim2020gan,salehi2020arae,akcay2018ganomaly}.

However, current contrastive learning-based approaches only consider the intuitively instance-level relationship among samples \cite{tack2020csi,li2021contrastive,zhong2021graph,han2021elsa}, and ignore other potential relations, such as the contrastive relations between samples and prototypes of normal samples,
and the discrimination between normal and abnormal samples.
The above relations are shown in Fig.~\ref{fig:representation}. 
Consequently, current contrastive learning-based approaches 
are prone to errors when 
distinguishing between normal and abnormal samples on the large contaminated set. Besides,
when contaminated with anomalies, a fine-tuning or an adaptation step \cite{sohn2020learning,han2021elsa} is usually required for obtaining better representations. These multi-stage training schemes often result in low efficiency with additional training tricks like early-stop strategy \cite{han2021elsa}, therefore not suitable for practice.

\begin{figure*}[!t]
\begin{center}
\includegraphics[width=0.85\linewidth]{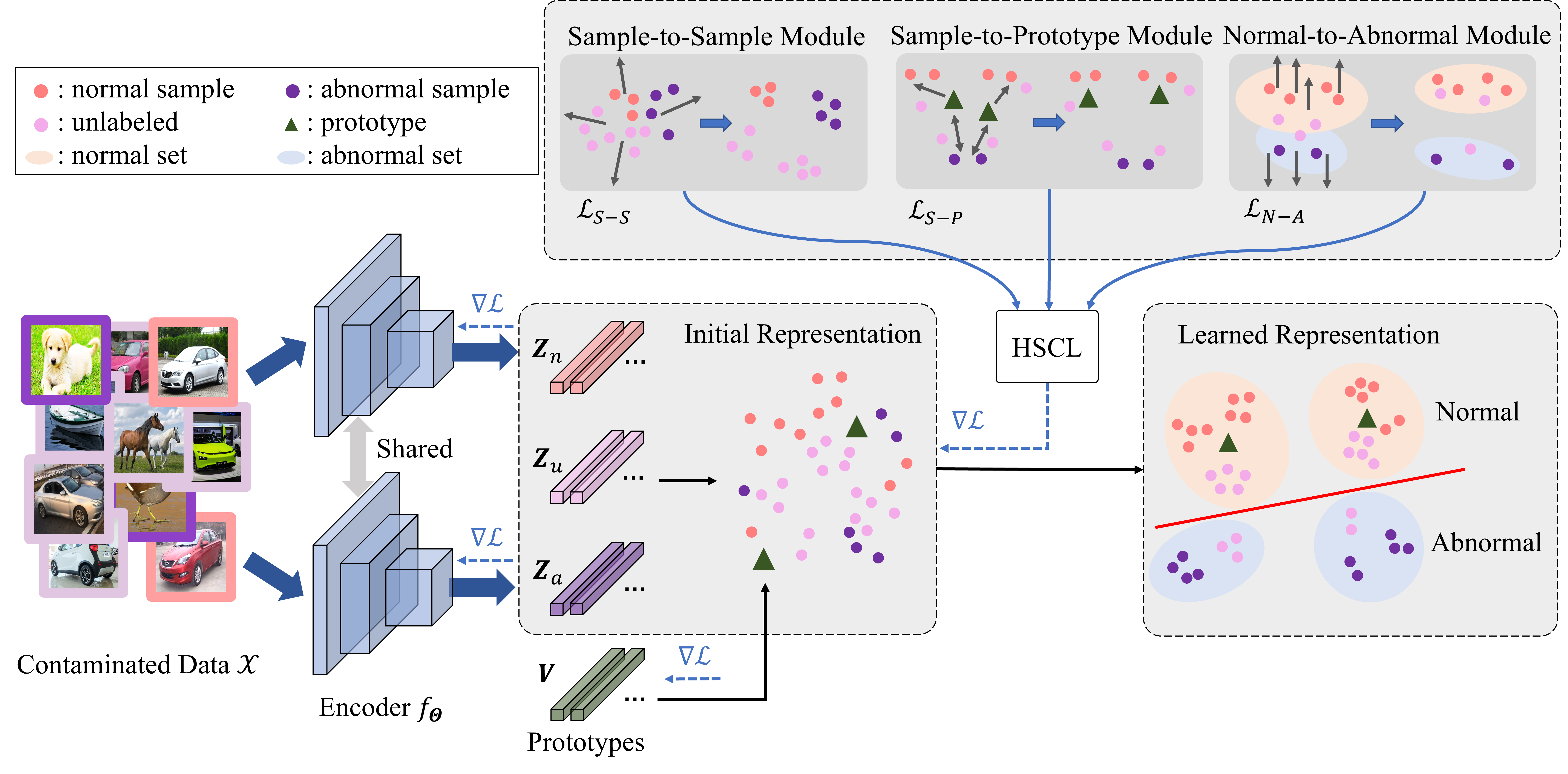}
\end{center}
   \caption{The framework of hierarchical semi-supervised contrastive learning for anomaly detection. Given a mixture of contaminated training data, three complementary sample-to-sample, sample-to-prototype, and normal-to-abnormal relations are learned in a hierarchical way. Specifically, the sample-to-sample relation is learned to enlarge the dissimilarities among different samples. Then, the prototypes are optimized for representing normal samples and pushing away anomalies to learn the sample-to-prototype relations. The normal-to-abnormal module further enlarges the discrimination between normal and abnormal samples.
   More details are demonstrated in Sec. \ref{sec:hier_CL}.}
\label{fig:flowchart}
\end{figure*}

To address the above issues, we propose a novel \underline{H}ierarchical \underline{S}emi-supervised \underline{C}ontrastive \underline{L}earning framework, termed HSCL, to identify anomalies with contaminated data under the semi-supervised setting. 
Specifically, HSCL jointly learns sample-to-sample, sample-to-prototype, and normal-to-abnormal relations to better distinguish anomalies over contaminated training data. 
The sample-to-sample relationship is learned following the basic InfoNCE loss \cite{oord2018representation} that enlarges the dissimilarities among different samples. 
Then, the similarities between prototypes and normal/abnormal samples are maximized/minimized, respectively, to regulate the sample-to-prototype contrastive relationship, where the prototypes \cite{lv2021learning} are defined as the representative features for normal samples.
Afterward, the soft weighting on the unlabeled samples is incorporated and is further used for sampling unlabeled data to learn the normal-to-abnormal relationship along with the labeled set. 
The framework is shown in Fig.~\ref{fig:flowchart}.
With the proposed hierarchical contrastive relations, HSCL achieves 1) end-to-end learning without offline clustering and fine-tuning that has high computational complexity, 2) discriminative learning from a limited number of labels, and 3) contaminated data mining from large unlabeled samples.
Extensive experiments are conducted under multiple contamination scenarios.
HSCL achieves state-of-the-art (SOTA) results on 1) CIFAR-10, CIFAR-100 \cite{krizhevsky2009learning}, and LSUN (FIX) \cite{liang2017enhancing,tack2020csi} for the one-class classification, and 2) ImageNet (FIX) \cite{hendrycks2019using,tack2020csi} and SVHN \cite{netzer2011reading} for cross-dataset anomaly detection.
Our main contributions are summarized as follows:
\begin{itemize}
\setlength{\itemsep}{0pt}
\setlength{\parsep}{0pt}
\setlength{\parskip}{0pt}
\item We present a novel end-to-end contamination-resistant anomaly detection framework using contaminated training data for discriminative representation learning.
\item We propose a hierarchical semi-supervised contrastive learning approach that jointly optimizes the complementary sample-to-sample, sample-to-prototype, and normal-to-abnormal relation in an online manner to enlarge the discrimination between normal and abnormal samples. 
\item We conduct extensive experiments with systematic analysis. The SOTA performance of HSCL on multiple scenarios and datasets validates the effectiveness of our HSCL. 
\end{itemize}

\section{Related Work}
\label{sec:related_work}

\paragraph{\textbf{Reconstruction-Based Anomaly Detection.}}
Reconstruction-based approaches assume that abnormal samples cannot be well represented and reconstructed with the model learned from clean normal data. The reconstruction error can be treated as an indicator of anomalies. The commonly used reconstruction-based techniques include the PCA methods \cite{kim2009observe}, sparse representation methods \cite{lu2013abnormal}, and recent auto-encoder-based methods \cite{zhou2017anomaly,zhao2017spatio,gong2019memorizing,huang2021esad}. For example, \cite{gong2019memorizing} introduces a memory module that can retrieve the most relevant memory items for reconstruction. However, studies have found that anomalies do not always yield a high reconstruction error when classes are similar \cite{zenati2018adversarially,somepalli2020unsupervised}.
Some studies employ generative adversarial networks (GAN) \cite{liu2018future,kim2020gan,salehi2020arae,akcay2018ganomaly} as the complement of the reconstruction loss. For example, \cite{liu2018future} adopts the U-Net \cite{ronneberger2015u} architecture and leverages the adversarial training to distinguish whether the predicted frame is real or fake after reconstruction. 
\cite{kim2020gan} proposes a novel GAN-based anomaly detection model, which consists of one auto-encoder generator and two separate discriminators for normal and anomalous inputs, respectively. However, it has been reported that GAN-based models easily generate suboptimal solutions and hence are inapplicable for complex datasets \cite{liu2020energy,pang2021deep}.

\paragraph{\textbf{Contrastive-Based Anomaly Detection.}}
The recent success of contrastive learning \cite{chen2020simple,he2020momentum,caron2020unsupervised,grill2020bootstrap,li2021contrastive,wang2022human} provides a potential manner for visual representations in anomaly detection \cite{tack2020csi,sohn2020learning,han2021elsa,reiss2021mean}. These contrastive learning-based approaches significantly outperform the conventional reconstruction-based approaches. For example, \cite{tack2020csi} proposes distributionally-shifted augmentations in contrastive learning, serving as a solid SOTA method in anomaly detection trained with clean normal data. A fine-tuning stage is employed to adapt the pre-trained features with mean-shifted contrastive loss in \cite{reiss2021mean}. To better address the contaminated data issue, \cite{han2021elsa} adopts a three-stage training scheme that fine-tunes the representations learned from contrastive loss with pseudo labels. 
However, existing contrastive learning-based approaches still suffer from the sensitivity to the contaminated abnormal samples or require multi-stage pre-training and fine-tuning, leading to low efficiency and suboptimal performance.

\paragraph{\textbf{Visual Representation Learning.}}
Recent works explore the visual representation learning \cite{kolesnikov2019revisiting,kolesnikov2020big,he2020momentum,wang2021dense} based on designing various self-supervised or pre-training tasks~\cite{Weihao22MetaFormer,Sucheng2022CVPR,yang2020CVPR,zhang2022vsa,xu2021vitae,zhan2020multi,zhan2018comprehensive,zhan2019exploring}, such as image inpainting \cite{pathak2016context}, permutation \cite{misra2020self}, predicting jigsaw puzzles \cite{kim2018learning}, and contrastive learning \cite{yuan2021multimodal,chen2020simple}. 
These learning strategies are also successfully extended to video representation learning, such as \cite{xu2019self,benaim2020speednet,tian2020contrastive,chen2021rspnet}. 
With much progress made recently, visual representation learning is employed in many real-world applications, such as anomaly detection \cite{zaheer2020claws,zaheer2022generative}, and human-based perception \cite{wang2021track,wang2022split,wang2022recent,wang2022human}. However, the complex hierarchical relationships among instances are seldom explored in the existing works.

\paragraph{\textbf{Semi-Supervised and Noisy Label Approaches.}}
Several approaches have been proposed for semi-supervised classification in recent years, such as MixMatch \cite{berthelot2019mixmatch}, EnAET \cite{wang2020enaet}, FixMatch \cite{sohn2020fixmatch}, SelfMatch \cite{kim2021selfmatch}, VPU \cite{chen2020variational} and ActiveMatch \cite{yuan2021activematch}. For example, FixMatch \cite{sohn2020fixmatch} uses the pseudo label generated from weakly augmented data to guide the prediction on strongly augmented data and achieves SOTA performance; VPU \cite{chen2020variational} learns from positive and unlabeled data.
Due to the effectiveness of semi-supervised approaches that take account of both the labeled set and the large unlabeled set,
many works have been proposed for anomaly detection under the semi-supervised setting \cite{gornitz2013toward,kingma2014semi,ruff2019deep,han2021elsa}. 
For example, \cite{ruff2019deep} modifies the one-class classifier that incorporates the negative abnormal samples in the training objective.
To deal with contaminated training data, we can treat anomaly detection as a noisy label problem. Since we assume that normal samples are dominant, we can regard all unlabeled samples as normal with noisy labels. Some progress has been made for noisy label classification, such as \cite{yi2019probabilistic,liu2020early,li2019dividemix,nishi2021augmentation}. For example, DivideMix \cite{li2019dividemix} leverages semi-supervised learning techniques with noisy labels. \cite{nishi2021augmentation} studies the effectiveness of several augmentation strategies.
However, these semi-supervised learning and noisy label approaches usually generate biased solutions when training data is imbalanced, particularly for the anomaly detection task where the normal data is always dominant.

\section{Method}

\subsection{Problem Description}
We consider the semi-supervised anomaly detection as the same as \cite{han2021elsa}. Let $\mathcal{X}=\{\mathbf{x}_i\}_{i=1}^N$ denote the training set, where $\mathcal{X}$ contains three disjoint sets, \textit{i.e.}, an unlabeled set $\mathcal{X}_u$, a labeled normal $\mathcal{X}_n$, and an abnormal set $\mathcal{X}_a$, respectively. Assume $\mathcal{X}_u$ is a contaminated set with a majority of normal samples and a small portion of abnormal samples. With the combination of these three sets $\mathcal{X}=\mathcal{X}_u \cup \mathcal{X}_n \cup \mathcal{X}_a$ in training, we aim at learning discriminative representations that can distinguish the anomalies in the testing data.

\subsection{Hierarchical Semi-Supervised Contrastive Learning}
\label{sec:hier_CL}

To detect anomalies with the contaminated training set, we propose a hierarchical semi-supervised contrastive learning approach that jointly learns sample-to-sample, sample-to-prototype, and normal-to-abnormal contrastive relations of instances, as shown in Fig.~\ref{fig:flowchart}. 
We demonstrate the details as follows.

\subsubsection{Sample-to-Sample Module}
The goal of this module is to learn the sample-to-sample relations and enlarge the dissimilarity among different samples. With only a few labeled samples, it is easy to get overfitting and collapse in trivial solutions. The learned representations of samples should include rich semantic information that can distinguish anomalies in unseen samples. To achieve this goal, we employ the InfoNCE-like loss \cite{oord2018representation,chen2020simple,tack2020csi} as follows,
\begin{equation}
\begin{aligned}
    & \mathcal{L}_{\text{InfoNCE}}(\mathbf{x}, \mathcal{X}_+, \mathcal{X}_-) \\ 
    & = -\frac{1}{N}\log{\frac{\sum_{\mathbf{x}^{'} \in \mathcal{X}_+}\exp{(\text{sim}(f_{\mathbf{\Theta}}(\mathbf{x}),f_{\mathbf{\Theta}}(\mathbf{x}^{'}))/\tau)}}{\sum_{\mathbf{x}^{'} \in (\mathcal{X}_+ \cup \mathcal{X}_-)}\exp{(\text{sim}(f_{\mathbf{\Theta}}(\mathbf{x}),f_{\mathbf{\Theta}}(\mathbf{x}^{'}))/\tau)}}},
\label{eq:InfoNCE}
\end{aligned}
\end{equation}
where $\mathcal{X}_+$ and $\mathcal{X}_-$ represent the set of positive and negative samples to $\mathbf{x}$, respectively; $``\text{sim}"$ stands for a similarity measure, \textit{e.g.}, cosine similarity; $\tau$ is the temperature parameter; and $f_{\mathbf{\Theta}}(\cdot)$ is the sample representation from the embedding network. We use the above InfoNCE loss as the training loss for the sample-to-sample module, \textit{i.e.}, $\mathcal{L}_{\text{S-S}}=\mathcal{L}_{\text{InfoNCE}}(\mathbf{x}, \mathcal{X}_+, \mathcal{X}_-)$. Following the typical framework of contrastive learning \cite{chen2020simple}, various augmented copies from the same instance are treated as positive samples, and copies from different instances are negative samples, \textit{i.e.},
\begin{equation}
\begin{aligned}
    & \mathcal{X}_+ = \{\mathbf{x}^{'}|\text{ID}(\mathbf{x}^{'})=\text{ID}(\mathbf{x})\}, \\
    & \mathcal{X}_- = \{\mathbf{x}^{'}|\text{ID}(\mathbf{x}^{'})\neq \text{ID}(\mathbf{x})\}.
\label{eq:id}
\end{aligned}
\end{equation}
Here, ID stands for the sample identity. In addition, following CSI \cite{tack2020csi}, we adopt the shifting transformation (rotation) to generate more out-of-distribution samples that are treated as negative copies of the original samples. 

\subsubsection{Sample-to-Prototype Module} 
The sample-to-sample module enlarges the dissimilarity among samples and learns neutral representations. However, using the sample-to-sample module alone lacks the discrimination on anomalies.
To this end, we propose a prototype learning scheme to learn discriminative representations with sample-to-prototype relations with the assistance of prototypes to represent normal samples. 
Unlike previous prototype-based approaches \cite{lv2021learning,han2021elsa} that use either the reconstruction constraint or offline clustering and omit the relations among labeled samples, we directly model the contrastive relations between samples and the prototypes.
As shown in Fig.~\ref{fig:flowchart}, we aim at generating prototypes that are close to normal samples and away from abnormal samples. The prototypes are differentiable and can be optimized in an online manner. With prototypes, it is easier to generate the sample weight for distinguishing abnormal samples, which will be demonstrated in Eq.~(\ref{eq:w}) and Eq.~(\ref{eq:sample}). Additionally, the learned prototypes are used as indicators in the inference stage to distinguish anomalies, as shown in Eq.~(\ref{eq:score}). The details of the sample-to-prototype module is explained as follows.

Denote the prototypes of normal samples as $\mathbf{V} \in \mathbb{R}^{D\times K}$, where $D$ is the feature dimension, and $K$ is the number of prototypes. Since the sample representation is updated for each batch data, $\mathbf{V}$ is also learned simultaneously in batch with the designed sample-to-prototype loss $\mathcal{L}_{\text{S-P}}$ defined as follows,
\begin{equation}
\begin{aligned}
    & \mathcal{L}_{\text{S-P}}(\mathcal{X}_n^{\tilde B} \cup \mathcal{X}_a^{\tilde B})
    = \frac{1}{N_n} \|\mathbf{b}_n-\text{max}_k(\mathbf{Z}_n^T\mathbf{V}_k)\|_2^2 + \frac{1}{N_a} \| [\text{max}_k(\mathbf{Z}_a^T\mathbf{V}_k) ]_+ \|_2^2,
\label{eq:Proto}
\end{aligned}
\end{equation}
where $\tilde B$ represents the augmented batch data. The subscripts ``\textit{n}'' and ``\textit{a}'' represent normal and abnormal samples, respectively. $\mathbf{Z}_n = [\mathbf{z}_1, \mathbf{z}_2, ..., \mathbf{z}_{N_n}] \in \mathbb{R}^{D\times N_n}$ are the normal sample embeddings, \textit{i.e.}, $\mathbf{z}_i=f_{\mathbf{\Theta}}(\mathbf{x}_i)$; $\mathbf{b}_n = [1, 1, ..., 1]^T$ is an all-one vector to represent the targeted similarity between normal samples to the closest prototype; $N_n = |\mathcal{X}_n^{\tilde B}|$ and $N_a = |\mathcal{X}_a^{\tilde B}|$ are the number of normal samples and abnormal samples in the augmented batch data, respectively. The loss function contains two parts. The first part pushes normal samples to be close to the learned prototypes $\mathbf{V}$, where $\text{max}(\cdot)$ takes the maximum similarity between each normal sample representation to the prototypes, constraining that normal samples should be close to at least one prototype. The second part is the constraint of the relationship between prototypes and abnormal samples, where $[\cdot]_+$ clamps negative values to zeros, thus pushing the prototypes to have non-positive cosine similarity scores to abnormal samples.
However, the above loss does not consider the large contaminated unlabeled samples. Instead, we consider the modified version with the sample weighting as follows,
\begin{equation}
\begin{aligned}
    & \mathcal{L}_{\text{S-P}}(\mathcal{X}_n^{\tilde B} \cup \mathcal{X}_u^{\tilde B} \cup \mathcal{X}_a^{\tilde B}) \\
    & = \frac{1}{\|\mathbf{w}\|_1} \|\mathbf{w}^T \left(\mathbf{b}_{n\cup u}-\text{max}_k\left(\mathbf{Z}_{n\cup u}^T\mathbf{V}_k\right)\right)\|_2^2 
    + \frac{1}{N_a} \| [\text{max}_k(\mathbf{Z}_a^T\mathbf{V}_k) ]_+ \|_2^2,
\label{eq:m_v2}
\end{aligned}
\end{equation}
where $\mathbf{Z}_{n\cup u}$ and $\mathbf{b}_{n\cup u}$ include both normal and unlabeled samples; and $\mathbf{w} \in \mathbb{R}^{N_{n\cup u}}$ is the sample weight defined as follows,
\begin{equation}
    \mathbf{w}_i = 
    \begin{cases}
    1, \ \ \ \ \ \ \ \ \ \ \ \ \ \ \ \ \ \ \ \ \ \ \ \ \ \ \ \ \ \ \ \ \ \ \ \ \text{if}\ \mathbf{x}_i \in \mathcal{X}_n;\\
    \left(\text{max}_k(\mathbf{z}_{i}^T \mathbf{V}_k) +1\right)/2, \  \ \ \ \ \ \ \ \ \text{if}\ \mathbf{x}_i \in \mathcal{X}_u.\\
    \end{cases}
\label{eq:w}
\end{equation}

We assume both $\mathbf{V}$ and $\mathbf{z}_i$ are already normalized. Therefore, $\mathbf{w}_i$ is in the range $[0, 1]$. If the representation of an unlabeled sample is closer to the learned prototypes, it is more likely to be a normal sample. Thus, the sample weight $\mathbf{w}_i$ is closer to 1. With the soft weighting strategy incorporated in the loss, the learned prototypes can better represent the normal samples.

\subsubsection{Normal-to-Abnormal Module}
As a complement to the previous two modules, the normal-to-abnormal module directly models the relation between normal and abnormal samples. Along with the assistance of the contamination-resistant sampling strategy, we aim at separating the abnormal representations from normal ones as much as possible.
To better utilize the unlabeled set $\mathcal{X}_u$ and the anomalies $\mathcal{X}_a$, we further employ the normal-to-abnormal contrastive relations in the training. To incorporate unlabeled data, we use sampling strategy and propose the normal-to-abnormal contrastive loss as follows, 
\begin{equation}
    \mathcal{L}_{\text{N-A}} = \mathcal{L}_{\text{InfoNCE}}( \mathbf{x}_{\mathbf{x}\sim p_\mathbf{w}}, \{\mathbf{x}^{'}\}_{\mathbf{x}^{'}\sim p_\mathbf{w}}, \mathcal{X}_{a}),
\label{eq:class-cl}
\end{equation}
where we treat abnormal samples $\mathcal{X}_{a}$ as negatives and draw positive samples from the distribution $p_\mathbf{w}$ defined by the sample weight $\mathbf{w}$, \textit{i.e.},
\begin{equation}
    p_{\mathbf{w}}(\mathbf{x}_i) = \frac{\mathbbm{1}_{[\mathbf{w}_i>\mathbf{w}_{\delta}]} \mathbf{w}_i}{\sum_i \mathbbm{1}_{[\mathbf{w}_i>\mathbf{w}_{\delta}]} \mathbf{w}_i},
\label{eq:sample}
\end{equation}
where the indicator function $\mathbbm{1}_{[\mathbf{w}_i>\mathbf{w}_{\delta}]}=1$ if and only if $\mathbf{w}_i$ is greater than a pre-defined threshold $\mathbf{w}_{\delta}$. This is to avoid sampling false positives as much as possible. 
Based on the definition of $\mathbf{w}$, normal samples are more likely to have larger weights than abnormal samples. Therefore, the normal instances have higher chances of being sampled.

\subsection{Training and Inference}
Combined with sample-to-sample, sample-to-prototype, and normal-to-abnormal hierarchical learning, the unified total loss is defined as follows,
\begin{equation}
    \mathcal{L} = \mathcal{L}_{\text{S-S}}+\lambda_1 \mathcal{L}_{\text{S-P}}+\lambda_2 \mathcal{L}_{\text{N-A}},
\label{eq:loss}
\end{equation}
where $\lambda_1$ and $\lambda_2$ are the weights to balance the contributions of different losses. 

In the inference stage, we define a normality score with the assistance of learned prototypes to distinguish anomalies as follows,
\begin{equation}
    s(\mathbf{\tilde x}_i|\mathbf{\hat \Theta}, \mathbf{\hat V}) = \text{max}_k\left(f_{\mathbf{\hat \Theta}}(\mathbf{\tilde x}_i)^T \mathbf{\hat V}_k\right),
\label{eq:score}
\end{equation}
where $\mathbf{\tilde x}_i$ represents a testing sample; $\mathbf{\hat \Theta}$ and $\mathbf{\hat V}$ represent the learned encoder parameters and prototypes, respectively. The normality score is measured as the maximum similarity between the testing sample and the learned prototypes. 

\section{Experiments}

To verify the effectiveness of our proposed HSCL, we consider three anomaly detection scenarios on several commonly used public datasets and compared HSCL with recent anomaly detection methods. Our HSCL significantly outperforms the general semi-supervised and noisy label approaches on the anomaly detection task with both quantitative and visualization results. Moreover, we also show the importance of each component of HSCL in the ablation study.

\begin{table}[t]
\centering
\caption{Experiment results of anomaly detection in Scenario-1 over CIFAR-10 with different labeled ratios $\gamma_l$. The best performance of each experiment is shown in bold.}
\begin{tabular}{L{2.8cm}|L{1.2cm}L{1.2cm}L{1.2cm}L{1.2cm}}
    \toprule
    $\gamma_l$ & .00 & .01 & .05 & .10 \\
    \midrule
    CSI \cite{tack2020csi} & \textbf{94.3} & - & - & - \\ 
    SS-DGM \cite{kingma2014semi} & - & 49.7 & 50.8 & 52.0 \\
    SSAD \cite{gornitz2013toward} & 62.0 & 73.0 & 71.5 & 70.1 \\
    DeepSAD \cite{ruff2019deep} & 62.9 & 72.6 & 77.9 & 79.8 \\
    Elsa \cite{han2021elsa} & - & 80.0 & 85.7 & 87.1 \\
    Elsa+ \cite{han2021elsa} & - & 94.3 & 95.2 & 95.5 \\
    \textbf{HSCL (Ours)} & - & \textbf{96.4} & \textbf{97.9} & \textbf{98.5} \\
    \bottomrule
\end{tabular}
\label{tab:s1}
\end{table}

\begin{table}[t]
\centering
\caption{Experiment results of anomaly detection in Scenario-2 over CIFAR-10 with different pollution ratios $\gamma_p$. The best performance of each experiment is shown in bold.}
\begin{tabular}{L{2.5cm}|L{1.cm}L{1.cm}L{1.cm}|L{2.5cm}|L{1.cm}L{1.cm}L{1.cm}}
    \toprule
    \multicolumn{3}{l}{Self-Supervised / Unsupervised} & & \multicolumn{3}{l}{Supervised / Semi-Supervised}\\
    \midrule
    $\gamma_p$ & .00 & .05 & .10 & $\gamma_p$ & .00 & .05 & .10\\
    \midrule
    OC-SVM \cite{scholkopf2001estimating} & 62.0 & 61.4 & 60.8 & SSAD \cite{gornitz2013toward} & 73.8 & 71.5 & 69.8 \\
    IF \cite{liu2008isolation} & 60.0 & 59.6 & 58.8 & SS-DGM \cite{kingma2014semi} & 50.8 & 50.1 & 50.5 \\
    KDE \cite{ruff2018deep} & 59.9 & 58.1 & 57.3 & DeepSAD \cite{ruff2019deep} & 77.9 & 74.0 & 71.8 \\
    DeepSVDD \cite{ruff2018deep} & 60.9 & 59.6 & 58.6 & Elsa \cite{han2021elsa} & 85.7 & 83.5 & 81.6 \\
    E$^3$Outlier \cite{wang2019effective} & 86.6 & 83.5 & 81.7 & Elsa+ \cite{han2021elsa} & 95.2 & 93.0 & 91.1 \\
    GOAD \cite{bergman2020classification} & 88.2 & 85.2 & 83.0 & \textbf{HSCL (Ours)} & \textbf{97.9} & \textbf{97.6} & \textbf{97.3} \\
    CSI \cite{tack2020csi} & 94.3 & 88.2 & 84.5 \\
    \bottomrule
\end{tabular}
\label{tab:s2}
\end{table}

\subsection{Scenario Setup}
In the experiments, we consider three representative scenarios following the prior work \cite{han2021elsa,akcay2018ganomaly,ruff2019deep,liu2020energy} with CIFAR-10 \cite{krizhevsky2009learning}, CIFAR-100 \cite{krizhevsky2009learning}, ImageNet (FIX) \cite{hendrycks2019using}, SVHN \cite{netzer2011reading}, and LSUN (FIX) \cite{liang2017enhancing} public datasets. Note that we use the fixed version of ImageNet and LSUN following the same process as mentioned in \cite{tack2020csi,han2021elsa}. The details of each scenario are described as follows.

\paragraph{\textbf{(Scenario-1) Semi-Supervised One-Class Classification}} \cite{han2021elsa,akcay2018ganomaly,ruff2019deep}.
We assume we can access a small subset of labeled normal samples $\mathcal{X}_n$ and abnormal samples $\mathcal{X}_a$ during training. We treat one class as the normal set while the remaining classes as anomalies. Both $\mathcal{X}_n$ and $\mathcal{X}_a$ are randomly sampled.
Denote the labeled ratio of $\mathcal{X}_n$ and $\mathcal{X}_a$ both as $\gamma_l$. We report the results on the testing set over 90 experiments (10 normal × 9 abnormal) for a given $\gamma_l$.

\paragraph{\textbf{(Scenario-2) Contaminated One-Class Classification}} \cite{han2021elsa,ruff2019deep}.
In this setting, in addition to a small labeled subset of normal and abnormal samples, we assume the normal training set is contaminated with anomalies with a pollution ratio $\gamma_p$. This is done by sampling images from every anomalous class and adding them into the unlabeled set $\mathcal{X}_u$. We report results with each pollution ratio $\gamma_p \in \{0.00, 0.05, 0.10\}$. 
$\gamma_l$ is fixed to 0.05 for all experiments.

\paragraph{\textbf{(Scenario-3) Cross-Dataset Anomaly Detection}} \cite{han2021elsa,liu2020energy}.
In this setting, we use all images in CIFAR-10 as normal samples and the down-sampled ImageNet dataset as labeled anomalies. We test the detection performance on the other four datasets, \textit{i.e.}, CIFAR-100, SVHN, LSUN, and ImageNet, as anomalies over the normal samples from the CIFAR-10 testing set. This setting tests the capability of the proposed method that leverages a large-scaled external dataset as an abnormal auxiliary set.

\paragraph{\textbf{Evaluation Metrics.}}
Following \cite{han2021elsa,tack2020csi}, we use the area under the receiver operating characteristics (AUROC) score as the evaluation metric. The ROC represents the true positive rate against the false-positive rate, while AUROC is the area under the curve. It is a common statistic for the goodness of a predictor in a binary classification task. The higher the score, the better the performance.

\subsection{Implementation Details}
\paragraph{\textbf{Training Details.}}
Following \cite{han2021elsa,tack2020csi}, we use ResNet-18 \cite{he2016deep} as the base encoder network and project images to representations with 128 dimensions. The temperature $\tau$ is set to 0.5 in the InfoNCE loss. $\lambda_1$ and $\lambda_2$ are set to 1. The number of prototypes $K$ is set to 1 for simplicity, and a more thorough analysis of the selection of $K$ is made in the ablation study. The batch size is set to 256. For the optimization, we train the proposed method with 250 epochs under Adam \cite{kingma2014adam} optimizer with an initial learning rate 1e-3. For the learning rate scheduler, we use the linear warmup \cite{goyal2017accurate} for the early 10 epochs, followed by the cosine decay schedule \cite{loshchilov2016sgdr}. The model is learned from scratch without any large-dataset pre-training. We use one Nvidia RTX 3090 GPU for training.

\paragraph{\textbf{Augmentation Details.}}
We use augmentations as the same with SimCLR \cite{chen2020simple}, including Inception crop \cite{szegedy2015going}, horizontal flip, color jitter, and gray-scale transform. We also adopt rotation with $\{90^{\circ}, 180^{\circ}, 270^{\circ}\}$ as shifting instances as defined in CSI \cite{tack2020csi}.

\begin{table}[t]
\centering
\caption{Experiment results of anomaly detection in Scenario-2 compared with general semi-supervised learning and noisy label approaches over CIFAR-10 (C-10), CIFAR-100 (C-100), and LSUN. The best performance of each experiment is shown in bold.}
\begin{tabular}{L{2.8cm}|C{1.0cm}C{1.0cm}|C{1.0cm}C{1.0cm}|C{1.0cm}C{1.0cm}}
    \toprule
    & \multicolumn{2}{c}{C-10} & \multicolumn{2}{c}{C-100}  & \multicolumn{2}{c}{LSUN} \\
    Method & .05 & .10 & .05 & .10 & .05 & .10 \\
    \midrule
    CSI \cite{tack2020csi} & 88.2 & 84.5 & 82.4 & 80.4 & 73.5 & 71.3 \\
    DivideMix \cite{li2019dividemix} & 83.9 & 83.2 & 66.8 & 66.3 & 67.8 & 66.9 \\
    Aug-LNL \cite{nishi2021augmentation} & 84.1 & 83.6 & 66.6 & 67.1 & 68.0 & 64.9\\
    FixMatch \cite{sohn2020fixmatch} & 93.5 & 94.3 & 71.8 & 78.5 & 76.0 & 73.2\\
    \textbf{HSCL (Ours)} & \textbf{97.6} & \textbf{97.3} & \textbf{93.0} & \textbf{92.2} & \textbf{88.6} & \textbf{88.9}\\
    \bottomrule
\end{tabular}

\label{tab:s2_semi}
\end{table}

\begin{table}[t]
\centering
\caption{Experiment results of anomaly detection in Scenario-3, where we use CIFAR-10 as in-distribution samples and other datasets as out-of-distribution samples. The best performance of each experiment is shown in bold type.}
\begin{tabular}{L{2.1cm}|L{1.3cm}L{1.3cm}L{1.3cm}L{1.3cm}}
    \toprule
    Dataset & GOAD & CSI & ELSA+ & \textbf{HSCL} \\
    \midrule
    ImageNet & 83.3 & 93.3 & 96.4 & \textbf{99.8}\\
    LSUN & 78.8 & 90.3 & 95.0 & \textbf{95.8}\\
    C-100 & 77.2 & \textbf{89.2} & 86.3 & 88.4\\
    SVHN & 96.3 & 99.8 & 99.4 & \textbf{99.8}\\
    \bottomrule
\end{tabular}
\label{tab:s3}
\end{table}

\subsection{Performance Comparison}

\paragraph{\textbf{Results on Scenario-1.}}
We report the AUROC score for scenario-1 in Table~\ref{tab:s1}. Several SOTA methods are adopted for comparison, including CSI \cite{tack2020csi}, SS-DGM \cite{kingma2014semi}, SSAD \cite{gornitz2013toward}, DeepSAD \cite{ruff2019deep}, and Elsa \cite{han2021elsa}. Here, CSI presents a novel detection method based on contrastive learning with shifting instances; SS-DGM proposes a semi-supervised generative model that allows for effective generalization from small labeled datasets to large
unlabeled ones; SSAD makes a detailed analysis of supervised anomaly detection with active learning strategy; DeepSAD presents
a deep end-to-end methodology for general semi-supervised anomaly detection method; and Elsa is a novel semi-supervised anomaly detection approach that unifies the concept of energy-based models
with contrastive learning. Except for CSI that belongs to self-supervised anomaly detection, the proposed HSCL achieves the best performance among all the compared semi-supervised methods, with more than 2\% improvement over the second-best method, Elsa+.

\paragraph{\textbf{Results on Scenario-2.}}
Different from Scenario-1 that we have clean labels for anomalies, we have a large unlabeled set contaminated with anomaly samples. In this scenario, we keep the labeled ratio fixed with $\gamma_l=0.05$ and conduct experiments with the contamination ratio $\gamma_p \in \{0.00, 0.05, 0.10\}$. The results are reported in Table~\ref{tab:s2}. Except for CSI, SS-DGM, SSAD, DeepSAD, and Elsa used in Scenario-1, we also compare with OC-SVM \cite{scholkopf2001estimating}, IF \cite{liu2008isolation}, KDE \cite{ruff2018deep}, DeepSVDD \cite{ruff2018deep}, E$^3$Outlier \cite{wang2019effective}, and GOAD \cite{bergman2020classification} methods. Specifically, OC-SVM is a method that estimates a function to distinguish from different distributions; IF is a fundamentally different model-based method that explicitly isolates anomalies instead of profiles normal points; DeepSVDD introduces a new anomaly detection method for one-class classification based on deep support vector data description, which is trained on an anomaly detection based objective; E$^3$Outlier first-time leverages a discriminative deep neural network for representation learning by using surrogate supervision to create multiple pseudo-classes from original data; and GOAD presents a unifying view and proposes an open-set method to relax generalization assumptions. From Table~\ref{tab:s2}, we can see our HSCL achieves the best performance, demonstrating the effectiveness of HSCL in semi-supervised settings. With the increase of $\gamma_p$, the performance degrades largely for most of the compared methods, while HSCL still roughly keeps the similar performance as the situation without contaminated samples. Compared with Elsa+, there is 2.7\%, 4.6\%, and 6.2\% significant improvement with $\gamma_p \in \{0.00, 0.05, 0.10\}$, respectively. This demonstrates the capability of contamination resistance of HSCL when dealing with a large number of unlabeled samples.

\begin{figure}[!t]
\begin{center}
\includegraphics[width=0.85\linewidth]{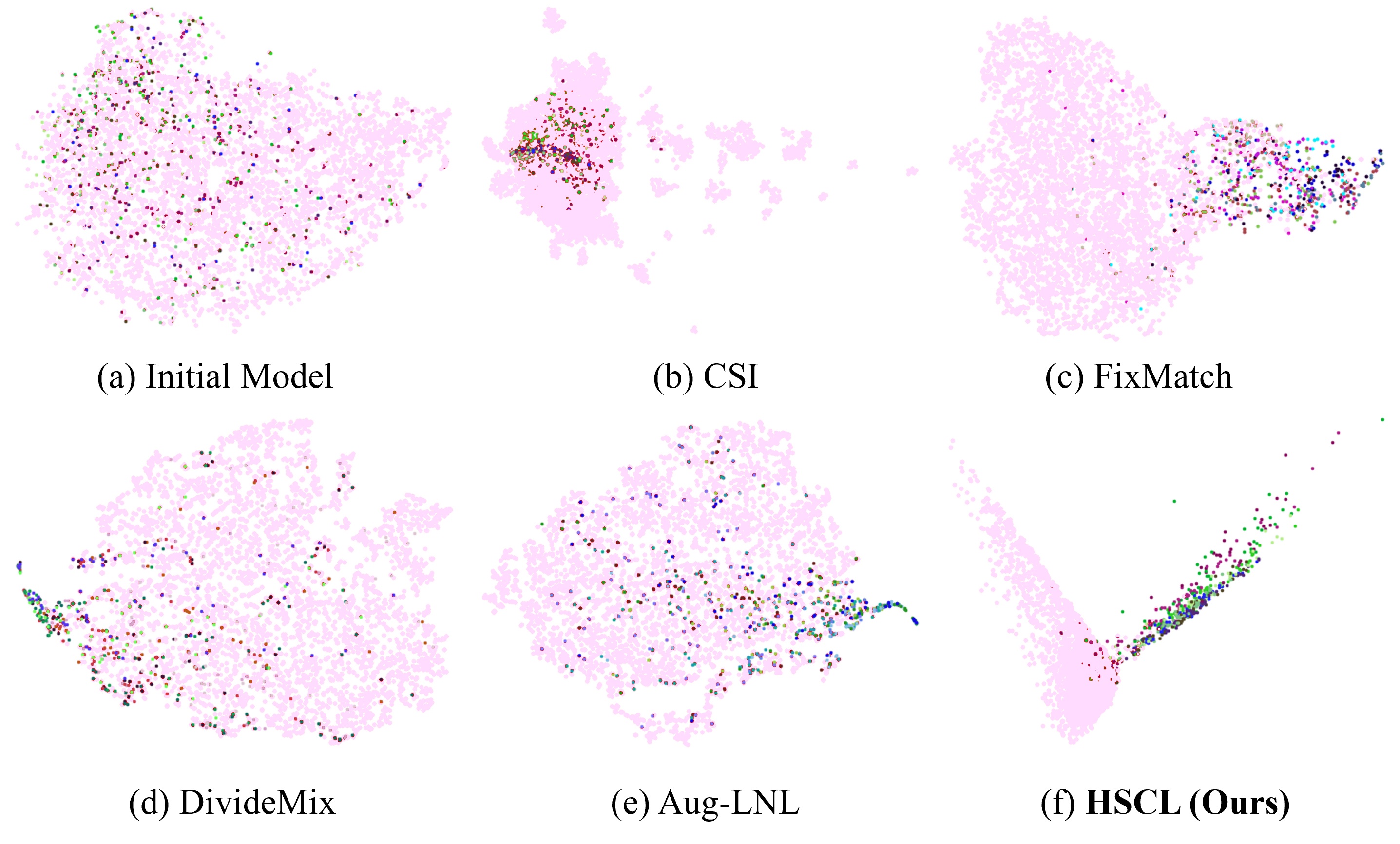}
\end{center}
   \caption{Visualization of learned representation for anomaly detection on CIFAR-10 using t-SNE. From (a) to (f): initial model, CSI, FixMatch, DivideMix, Aug-LNL, and HSCL. Normal samples (from class 0) are in pink color, while abnormal samples (from other classes) are in other colors.}
\label{fig:vis_2}
\end{figure}

\paragraph{\textbf{Comparison with General Semi-Supervised and Noisy Label Approaches.}}
The contaminated scenario can be treated as a special case of the general semi-supervised learning setting. The main difference between contaminated anomaly detection and general semi-supervised learning is that the normal sample is dominant in the unlabeled set in anomaly detection, which is an imbalanced data problem. We compare one of the SOTA methods, FixMatch \cite{sohn2020fixmatch}, under the anomaly detection setting and report the results on CIFAR-10, CIFAR-100, and LSUN datasets in Table~\ref{tab:s2_semi}. Our proposed HSCL outperforms FixMatch with a large margin, validating the effectiveness of our method for anomaly detection.
Note that FixMatch achieves higher performance with $\gamma_p = .10$ than $\gamma_p = .05$ on CIFAR-10 and CIFAR-100. This is reasonable since using $\gamma_p = .05$ suffers from a more severe imbalanced problem, resulting in a larger degeneration on the performance. 
In addition, the contaminated setting can also be treated as a special case of noisy-label classification. We assume all samples from the contaminated set are from the normal class with some noisy labels. We also compare our proposed HSCL with two recent SOTA methods dealing with noisy label classification, \textit{i.e.}, DivideMix \cite{li2019dividemix} and Aug-LNL \cite{nishi2021augmentation}. As shown in Table~\ref{tab:s2_semi}, HSCL also outperforms both the methods on CIFAR-10, CIFAR-100, and LSUN datasets. 

\paragraph{\textbf{Results on Scenario-3.}}
The cross-dataset validation is shown in Table~\ref{tab:s3}, where we treat CIFAR-10 as the normal set, while four other datasets, \textit{i.e.}, ImageNet, LSUN, CIFAR-100, and SVHN, as the abnormal set. HSCL outperforms other methods like GOAD, CSI, and Elsa+ in most of the cases. This further verifies the generalization of the proposed HSCL method.

\paragraph{\textbf{Visualizations.}}
To verify the effectiveness of the proposed hierarchical contrastive learning strategy, We draw the sample representations on CIFAR-10 with t-SNE as the dimension reduction method and compare with the initial model, CSI, FixMatch, DivideMix, and Aug-LNL, as shown in Fig.~\ref{fig:vis_2}. All the models are trained using mixed contaminated data with $\gamma_l=0.05$ and $\gamma_p=0.05$. In this example, the \textit{plane} class is treated as the normal class while samples from other classes are regarded as anomalies. In each sub-figure, we use light pink color to represent the normal class while other colors represent abnormal classes. As shown in the figure, the anomalies can be better distinguished using HSCL compared with other methods.

\subsection{Ablation Study}

\begin{table}[t]
\centering
\caption{Results with different settings of module components on C-10, C-100 (short for CIFAR-10 and CIFAR-100) and LSUN datasets. ``S-S", ``S-P" and ``N-A" represent sample-to-sample, sample-to-prototype, and normal-to-abnormal, respectively. ``w/o pos in S-P" and ``w/o neg in S-P" represent the performance without the first and second term in Eq. (\ref{eq:m_v2}), respectively.}
\begin{tabular}{L{2.8cm}|L{1.3cm}L{1.3cm}L{1.3cm}}
    \toprule
    Component & C-10 & C-100 & LSUN \\
    \midrule
    w/o S-S & 89.8 & 72.9 & 67.0 \\
    w/o S-P & 82.9 & 85.5 & 70.3 \\ 
    w/o N-A & 94.1 & 85.3 & 87.1 \\
    w/o pos in S-P & 96.5 & 88.4 & 87.3 \\
    w/o neg in S-P & 96.1 & 89.4 & 87.8 \\
    Full model & \textbf{97.1} & \textbf{90.5} & \textbf{88.6} \\
    \bottomrule
\end{tabular}
\label{tab:module}
\end{table}

\begin{figure}[!t]
\begin{center}
\includegraphics[width=0.9\linewidth]{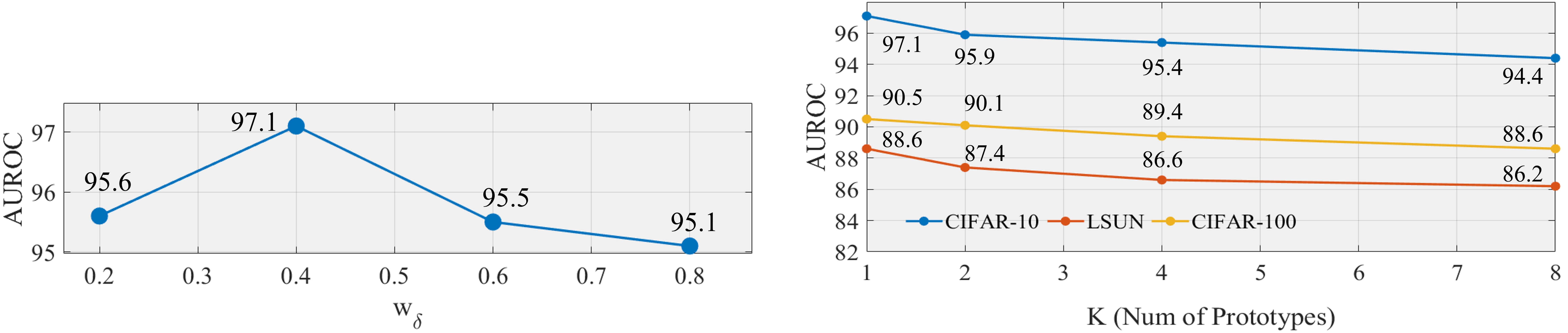}
\end{center}
   \caption{Left: AUROC with variant sampling threshold $\mathbf{w}_{\delta}$ on CIFAR-10. Right: AUROC with variant number of prototypes on CIFAR-10, CIFAR-100 and LSUN datasets.}
\label{fig:w_thres}
\end{figure}

\paragraph{\textbf{Importance of Different Modules.}}
To validate the effect of individual modules, we conduct experiments with the removal of each module. For simplicity, we use ``S-S", ``S-P", and ``N-A" to represent sample-to-sample, sample-to-prototype, and normal-to-abnormal modules, respectively. Note that we use k-NN to distinguish anomalies as the same as \cite{tack2020csi} when removing the sample-to-prototype module since prototypes are not available at this configuration. We further verify the importance of individual terms of the prototype learning in Eq.~(\ref{eq:m_v2}). We report the results on CIFAR-10, CIFAR-100, and LSUN datasets, as shown in Table \ref{tab:module}. As expected, the full model generates the best performance. The drop of ``w/o S-P" is due to two facts: 1) prototypes are replaced with k-NN in the normality score and 2) the soft-weighting strategy has to be discarded since prototypes are unavailable. The above changes make the method sensitive to unlabeled abnormal samples, causing the degradation. Prototype learning remains even though we discard only one term in Eq.~(\ref{eq:m_v2}). That is why the degradation of ``w/o pos in S-P" and ``w/o neg in S-P" is not significant.

\paragraph{\textbf{Varying Sampling Threshold.}}
To optimize the normal-to-abnormal contrastive learning loss, we sample the data from the unlabeled set according to the sampling strategy with a pre-defined weight threshold $\mathbf{w}_{\delta}$ in Eq.~(\ref{eq:sample}). To learn the effect and sensitivity of the threshold, we vary $\mathbf{w}_{\delta}$ from 0.2 to 0.8 and report the AUROC of \textit{plane} class on CIFAR-10 dataset in the left of Fig.~\ref{fig:w_thres}. From the result, when $\mathbf{w}_{\delta}=0.4$, we achieve the best performance with $\text{AUROC}=97.1\%$. With different choices of thresholds, the result does not change much, showing the robustness of the sampling strategy.

\paragraph{\textbf{Analysis of Prototypes.}}
We vary the prototype number $K$ from 1 to 8 on CIFAR-10, CIFAR-100, and LSUN datasets. The results of AUROC are shown in the right of Fig.~\ref{fig:w_thres}. We achieve the best performance with $K=1$. With the increase of $K$, the performance slightly degrades. 
This phenomenon is reasonable for the contamination setting. 
Since the learned prototypes need to represent normal samples (not abnormal samples), using one prototype to represent the normal class is good enough. However, when we increase the number of prototypes, some prototypes may become closer to unlabeled abnormal samples in the training set, resulting in slightly poorer performance for anomaly detection.

\section{Conclusion}
In this paper, we tackle anomaly detection in a semi-supervised setting with contaminated samples. We learn the discriminative sample representations based on hierarchical contrastive relations among samples, prototypes, and classes in an end-to-end manner. The proposed method, HSCL, achieves the SOTA performance in several different scenarios and outperforms recent general semi-supervised learning and noisy-label approaches. Furthermore, the ablation study, including the analysis of components of individual modules, the sampling strategy with soft weighting, and the number of prototypes, also demonstrates the effectiveness and robustness of the proposed method. In our future work, we plan to conduct anomaly detection with contaminated data in unsupervised settings. 

\paragraph{\textbf{Acknowledgements.}}
This work is supported by National Natural Science Foundation of China (62106219, U20B2066, 61976186),  
the Fundamental Research Funds for the Central Universities (226-2022-00087),
NUS Advanced Research and Technology Innovation Centre (Project Reference: ECT-RP2),
and NUS  Faculty Research Committee Grant (WBS: A-0009440-00-00).



\clearpage
%
%
\bibliographystyle{splncs04}
\bibliography{egbib}

\begin{thebibliography}{10}
\providecommand{\url}[1]{\texttt{#1}}
\providecommand{\urlprefix}{URL }
\providecommand{\doi}[1]{https://doi.org/#1}

\bibitem{akcay2018ganomaly}
Akcay, S., Atapour-Abarghouei, A., Breckon, T.P.: Ganomaly: Semi-supervised
  anomaly detection via adversarial training. In: Asian conference on computer
  vision. pp. 622--637. Springer (2018)

\bibitem{benaim2020speednet}
Benaim, S., Ephrat, A., Lang, O., Mosseri, I., Freeman, W.T., Rubinstein, M.,
  Irani, M., Dekel, T.: Speednet: Learning the speediness in videos. In:
  Proceedings of the IEEE Conference on Computer Vision and Pattern
  Recognition. pp. 9922--9931 (2020)

\bibitem{bergman2020classification}
Bergman, L., Hoshen, Y.: Classification-based anomaly detection for general
  data. arXiv preprint arXiv:2005.02359  (2020)

\bibitem{berthelot2019mixmatch}
Berthelot, D., Carlini, N., Goodfellow, I., Papernot, N., Oliver, A., Raffel,
  C.: Mixmatch: A holistic approach to semi-supervised learning. arXiv preprint
  arXiv:1905.02249  (2019)

\bibitem{caron2020unsupervised}
Caron, M., Misra, I., Mairal, J., Goyal, P., Bojanowski, P., Joulin, A.:
  Unsupervised learning of visual features by contrasting cluster assignments.
  arXiv preprint arXiv:2006.09882  (2020)

\bibitem{chen2020variational}
Chen, H., Liu, F., Wang, Y., Zhao, L., Wu, H.: A variational approach for
  learning from positive and unlabeled data. Advances in Neural Information
  Processing Systems  \textbf{33},  14844--14854 (2020)

\bibitem{chen2021rspnet}
Chen, P., Huang, D., He, D., Long, X., Zeng, R., Wen, S., Tan, M., Gan, C.:
  Rspnet: Relative speed perception for unsupervised video representation
  learning. In: Proceedings of the AAAI Conference on Artificial Intelligence.
  vol.~35, pp. 1045--1053 (2021)

\bibitem{chen2020simple}
Chen, T., Kornblith, S., Norouzi, M., Hinton, G.: A simple framework for
  contrastive learning of visual representations. In: International conference
  on machine learning. pp. 1597--1607. PMLR (2020)

\bibitem{chen2021unsupervised}
Chen, Y., Tian, Y., Pang, G., Carneiro, G.: Unsupervised anomaly detection with
  multi-scale interpolated gaussian descriptors. arXiv preprint
  arXiv:2101.10043  (2021)

\bibitem{gong2019memorizing}
Gong, D., Liu, L., Le, V., Saha, B., Mansour, M.R., Venkatesh, S., Hengel,
  A.v.d.: Memorizing normality to detect anomaly: Memory-augmented deep
  autoencoder for unsupervised anomaly detection. In: Proceedings of the
  IEEE/CVF International Conference on Computer Vision. pp. 1705--1714 (2019)

\bibitem{gornitz2013toward}
G{\"o}rnitz, N., Kloft, M., Rieck, K., Brefeld, U.: Toward supervised anomaly
  detection. Journal of Artificial Intelligence Research  \textbf{46},
  235--262 (2013)

\bibitem{goyal2017accurate}
Goyal, P., Doll{\'a}r, P., Girshick, R., Noordhuis, P., Wesolowski, L., Kyrola,
  A., Tulloch, A., Jia, Y., He, K.: Accurate, large minibatch sgd: Training
  imagenet in 1 hour. arXiv preprint arXiv:1706.02677  (2017)

\bibitem{grill2020bootstrap}
Grill, J.B., Strub, F., Altch{\'e}, F., Tallec, C., Richemond, P.H.,
  Buchatskaya, E., Doersch, C., Pires, B.A., Guo, Z.D., Azar, M.G., et~al.:
  Bootstrap your own latent: A new approach to self-supervised learning. arXiv
  preprint arXiv:2006.07733  (2020)

\bibitem{han2021elsa}
Han, S., Song, H., Lee, S., Park, S., Cha, M.: Elsa: Energy-based learning for
  semi-supervised anomaly detection. arXiv preprint arXiv:2103.15296  (2021)

\bibitem{he2020momentum}
He, K., Fan, H., Wu, Y., Xie, S., Girshick, R.: Momentum contrast for
  unsupervised visual representation learning. In: Proceedings of the IEEE/CVF
  Conference on Computer Vision and Pattern Recognition. pp. 9729--9738 (2020)

\bibitem{he2016deep}
He, K., Zhang, X., Ren, S., Sun, J.: Deep residual learning for image
  recognition. In: Proceedings of the IEEE conference on computer vision and
  pattern recognition. pp. 770--778 (2016)

\bibitem{hendrycks2019using}
Hendrycks, D., Mazeika, M., Kadavath, S., Song, D.: Using self-supervised
  learning can improve model robustness and uncertainty. arXiv preprint
  arXiv:1906.12340  (2019)

\bibitem{kim2021selfmatch}
Kim, B., Choo, J., Kwon, Y.D., Joe, S., Min, S., Gwon, Y.: Selfmatch: Combining
  contrastive self-supervision and consistency for semi-supervised learning.
  arXiv preprint arXiv:2101.06480  (2021)

\bibitem{kim2018learning}
Kim, D., Cho, D., Yoo, D., Kweon, I.S.: Learning image representations by
  completing damaged jigsaw puzzles. In: 2018 IEEE Winter Conference on
  Applications of Computer Vision. pp. 793--802. IEEE (2018)

\bibitem{kim2009observe}
Kim, J., Grauman, K.: Observe locally, infer globally: a space-time mrf for
  detecting abnormal activities with incremental updates. In: 2009 IEEE
  conference on computer vision and pattern recognition. pp. 2921--2928. IEEE
  (2009)

\bibitem{kim2020gan}
Kim, J., Jeong, K., Choi, H., Seo, K.: Gan-based anomaly detection in imbalance
  problems. In: European Conference on Computer Vision. pp. 128--145. Springer
  (2020)

\bibitem{kingma2014adam}
Kingma, D.P., Ba, J.: Adam: A method for stochastic optimization. arXiv
  preprint arXiv:1412.6980  (2014)

\bibitem{kingma2014semi}
Kingma, D.P., Mohamed, S., Rezende, D.J., Welling, M.: Semi-supervised learning
  with deep generative models. In: Advances in neural information processing
  systems. pp. 3581--3589 (2014)

\bibitem{kolesnikov2020big}
Kolesnikov, A., Beyer, L., Zhai, X., Puigcerver, J., Yung, J., Gelly, S.,
  Houlsby, N.: Big transfer (bit): General visual representation learning. In:
  European conference on computer vision. pp. 491--507. Springer (2020)

\bibitem{kolesnikov2019revisiting}
Kolesnikov, A., Zhai, X., Beyer, L.: Revisiting self-supervised visual
  representation learning. In: Proceedings of the IEEE/CVF conference on
  computer vision and pattern recognition. pp. 1920--1929 (2019)

\bibitem{krizhevsky2009learning}
Krizhevsky, A., Hinton, G., et~al.: Learning multiple layers of features from
  tiny images  (2009)

\bibitem{li2021cutpaste}
Li, C.L., Sohn, K., Yoon, J., Pfister, T.: Cutpaste: Self-supervised learning
  for anomaly detection and localization. In: Proceedings of the IEEE/CVF
  Conference on Computer Vision and Pattern Recognition. pp. 9664--9674 (2021)

\bibitem{li2019dividemix}
Li, J., Socher, R., Hoi, S.C.: Dividemix: Learning with noisy labels as
  semi-supervised learning. In: International Conference on Learning
  Representations (2019)

\bibitem{li2021contrastive}
Li, Y., Hu, P., Liu, Z., Peng, D., Zhou, J.T., Peng, X.: Contrastive
  clustering. In: 2021 AAAI Conference on Artificial Intelligence (AAAI) (2021)

\bibitem{liang2017enhancing}
Liang, S., Li, Y., Srikant, R.: Enhancing the reliability of
  out-of-distribution image detection in neural networks. arXiv preprint
  arXiv:1706.02690  (2017)

\bibitem{liu2008isolation}
Liu, F.T., Ting, K.M., Zhou, Z.H.: Isolation forest. In: 2008 eighth ieee
  international conference on data mining. pp. 413--422. IEEE (2008)

\bibitem{liu2020early}
Liu, S., Niles-Weed, J., Razavian, N., Fernandez-Granda, C.: Early-learning
  regularization prevents memorization of noisy labels. Advances in Neural
  Information Processing Systems  \textbf{33} (2020)

\bibitem{liu2020energy}
Liu, W., Wang, X., Owens, J.D., Li, Y.: Energy-based out-of-distribution
  detection. arXiv preprint arXiv:2010.03759  (2020)

\bibitem{liu2018future}
Liu, W., Luo, W., Lian, D., Gao, S.: Future frame prediction for anomaly
  detection--a new baseline. In: Proceedings of the IEEE conference on computer
  vision and pattern recognition. pp. 6536--6545 (2018)

\bibitem{loshchilov2016sgdr}
Loshchilov, I., Hutter, F.: Sgdr: Stochastic gradient descent with warm
  restarts. arXiv preprint arXiv:1608.03983  (2016)

\bibitem{lu2013abnormal}
Lu, C., Shi, J., Jia, J.: Abnormal event detection at 150 fps in matlab. In:
  Proceedings of the IEEE international conference on computer vision. pp.
  2720--2727 (2013)

\bibitem{lv2021learning}
Lv, H., Chen, C., Cui, Z., Xu, C., Li, Y., Yang, J.: Learning normal dynamics
  in videos with meta prototype network. In: Proceedings of the IEEE/CVF
  Conference on Computer Vision and Pattern Recognition. pp. 15425--15434
  (2021)

\bibitem{ma2022deep}
Ma, R., Pang, G., Chen, L., van~den Hengel, A.: Deep graph-level anomaly
  detection by glocal knowledge distillation. In: Proceedings of the Fifteenth
  ACM International Conference on Web Search and Data Mining. pp. 704--714
  (2022)

\bibitem{misra2020self}
Misra, I., Maaten, L.v.d.: Self-supervised learning of pretext-invariant
  representations. In: Proceedings of the IEEE Conference on Computer Vision
  and Pattern Recognition. pp. 6707--6717 (2020)

\bibitem{netzer2011reading}
Netzer, Y., Wang, T., Coates, A., Bissacco, A., Wu, B., Ng, A.Y.: Reading
  digits in natural images with unsupervised feature learning  (2011)

\bibitem{nishi2021augmentation}
Nishi, K., Ding, Y., Rich, A., Hollerer, T.: Augmentation strategies for
  learning with noisy labels. In: Proceedings of the IEEE/CVF Conference on
  Computer Vision and Pattern Recognition. pp. 8022--8031 (2021)

\bibitem{oord2018representation}
Oord, A.v.d., Li, Y., Vinyals, O.: Representation learning with contrastive
  predictive coding. arXiv preprint arXiv:1807.03748  (2018)

\bibitem{pang2021explainable}
Pang, G., Ding, C., Shen, C., Hengel, A.v.d.: Explainable deep few-shot anomaly
  detection with deviation networks. arXiv preprint arXiv:2108.00462  (2021)

\bibitem{pang2021toward}
Pang, G., van~den Hengel, A., Shen, C., Cao, L.: Toward deep supervised anomaly
  detection: Reinforcement learning from partially labeled anomaly data. In:
  Proceedings of the 27th ACM SIGKDD conference on knowledge discovery \& data
  mining. pp. 1298--1308 (2021)

\bibitem{pang2021deep}
Pang, G., Shen, C., Cao, L., Hengel, A.V.D.: Deep learning for anomaly
  detection: A review. ACM Computing Surveys (CSUR)  \textbf{54}(2),  1--38
  (2021)

\bibitem{pathak2016context}
Pathak, D., Krahenbuhl, P., Donahue, J., Darrell, T., Efros, A.A.: Context
  encoders: Feature learning by inpainting. In: Proceedings of the IEEE
  Conference on Computer Vision and Pattern Recognition. pp. 2536--2544 (2016)

\bibitem{qiu2022latent}
Qiu, C., Li, A., Kloft, M., Rudolph, M., Mandt, S.: Latent outlier exposure for
  anomaly detection with contaminated data. arXiv preprint arXiv:2202.08088
  (2022)

\bibitem{reiss2021mean}
Reiss, T., Hoshen, Y.: Mean-shifted contrastive loss for anomaly detection.
  arXiv preprint arXiv:2106.03844  (2021)

\bibitem{Sucheng2022CVPR}
Ren, S., Zhou, D., He, S., Feng, J., Wang, X.: Shunted self-attention via
  multi-scale token aggregation. In: Proceedings of the IEEE/CVF Conference on
  Computer Vision and Pattern Recognition (2022)

\bibitem{ronneberger2015u}
Ronneberger, O., Fischer, P., Brox, T.: U-net: Convolutional networks for
  biomedical image segmentation. In: International Conference on Medical image
  computing and computer-assisted intervention. pp. 234--241. Springer (2015)

\bibitem{ruff2018deep}
Ruff, L., Vandermeulen, R., Goernitz, N., Deecke, L., Siddiqui, S.A., Binder,
  A., M{\"u}ller, E., Kloft, M.: Deep one-class classification. In:
  International conference on machine learning. pp. 4393--4402. PMLR (2018)

\bibitem{ruff2019deep}
Ruff, L., Vandermeulen, R.A., G{\"o}rnitz, N., Binder, A., M{\"u}ller, E.,
  M{\"u}ller, K.R., Kloft, M.: Deep semi-supervised anomaly detection. arXiv
  preprint arXiv:1906.02694  (2019)

\bibitem{salehi2020arae}
Salehi, M., Arya, A., Pajoum, B., Otoofi, M., Shaeiri, A., Rohban, M.H.,
  Rabiee, H.R.: Arae: Adversarially robust training of autoencoders improves
  novelty detection. arXiv preprint arXiv:2003.05669  (2020)

\bibitem{scholkopf2001estimating}
Sch{\"o}lkopf, B., Platt, J.C., Shawe-Taylor, J., Smola, A.J., Williamson,
  R.C.: Estimating the support of a high-dimensional distribution. Neural
  computation  \textbf{13}(7),  1443--1471 (2001)

\bibitem{sohn2020fixmatch}
Sohn, K., Berthelot, D., Carlini, N., Zhang, Z., Zhang, H., Raffel, C.A.,
  Cubuk, E.D., Kurakin, A., Li, C.L.: Fixmatch: Simplifying semi-supervised
  learning with consistency and confidence. Advances in neural information
  processing systems  \textbf{33},  596--608 (2020)

\bibitem{sohn2020learning}
Sohn, K., Li, C.L., Yoon, J., Jin, M., Pfister, T.: Learning and evaluating
  representations for deep one-class classification. arXiv preprint
  arXiv:2011.02578  (2020)

\bibitem{somepalli2020unsupervised}
Somepalli, G., Wu, Y., Balaji, Y., Vinzamuri, B., Feizi, S.: Unsupervised
  anomaly detection with adversarial mirrored autoencoders. arXiv preprint
  arXiv:2003.10713  (2020)

\bibitem{szegedy2015going}
Szegedy, C., Liu, W., Jia, Y., Sermanet, P., Reed, S., Anguelov, D., Erhan, D.,
  Vanhoucke, V., Rabinovich, A.: Going deeper with convolutions. In:
  Proceedings of the IEEE conference on computer vision and pattern
  recognition. pp.~1--9 (2015)

\bibitem{tack2020csi}
Tack, J., Mo, S., Jeong, J., Shin, J.: Csi: Novelty detection via contrastive
  learning on distributionally shifted instances. Advances in neural
  information processing systems  \textbf{33},  11839--11852 (2020)

\bibitem{tian2020contrastive}
Tian, Y., Krishnan, D., Isola, P.: Contrastive multiview coding. In: European
  Conference on Computer Vision. pp. 776--794. Springer (2020)

\bibitem{wang2021track}
Wang, G., Gu, R., Liu, Z., Hu, W., Song, M., Hwang, J.N.: Track without
  appearance: Learn box and tracklet embedding with local and global motion
  patterns for vehicle tracking. In: Proceedings of the IEEE/CVF International
  Conference on Computer Vision. pp. 9876--9886 (2021)

\bibitem{wang2022recent}
Wang, G., Song, M., Hwang, J.N.: Recent advances in embedding methods for
  multi-object tracking: A survey. arXiv preprint arXiv:2205.10766  (2022)

\bibitem{wang2022split}
Wang, G., Wang, Y., Gu, R., Hu, W., Hwang, J.N.: Split and connect: A universal
  tracklet booster for multi-object tracking. IEEE Transactions on Multimedia
  (2022)

\bibitem{wang2019anomaly}
Wang, G., Yuan, X., Zheng, A., Hsu, H.M., Hwang, J.N.: Anomaly candidate
  identification and starting time estimation of vehicles from traffic videos.
  In: CVPR Workshops. pp. 382--390 (2019)

\bibitem{wang2022human}
Wang, G., Lu, K., Zhou, Y., He, Z., Wang, G.: Human-centered prior-guided and
  task-dependent multi-task representation learning for action recognition
  pre-training. arXiv preprint arXiv:2204.12729  (2022)

\bibitem{wang2019effective}
Wang, S., Zeng, Y., Liu, X., Zhu, E., Yin, J., Xu, C., Kloft, M.: Effective
  end-to-end unsupervised outlier detection via inlier priority of
  discriminative network  (2019)

\bibitem{wang2020enaet}
Wang, X., Kihara, D., Luo, J., Qi, G.J.: Enaet: A self-trained framework for
  semi-supervised and supervised learning with ensemble transformations. IEEE
  Transactions on Image Processing  \textbf{30},  1639--1647 (2020)

\bibitem{wang2021dense}
Wang, X., Zhang, R., Shen, C., Kong, T., Li, L.: Dense contrastive learning for
  self-supervised visual pre-training. In: Proceedings of the IEEE/CVF
  Conference on Computer Vision and Pattern Recognition. pp. 3024--3033 (2021)

\bibitem{xu2019self}
Xu, D., Xiao, J., Zhao, Z., Shao, J., Xie, D., Zhuang, Y.: Self-supervised
  spatiotemporal learning via video clip order prediction. In: Proceedings of
  the IEEE Conference on Computer Vision and Pattern Recognition. pp.
  10334--10343 (2019)

\bibitem{xu2021vitae}
Xu, Y., Zhang, Q., Zhang, J., Tao, D.: Vitae: Vision transformer advanced by
  exploring intrinsic inductive bias. Advances in Neural Information Processing
  Systems  \textbf{34} (2021)

\bibitem{yang2020CVPR}
Yang, Y., Qiu, J., Song, M., Tao, D., Wang, X.: Distilling knowledge from graph
  convolutional networks. In: Proceedings of the IEEE/CVF Conference on
  Computer Vision and Pattern Recognition (2020)

\bibitem{yi2019probabilistic}
Yi, K., Wu, J.: Probabilistic end-to-end noise correction for learning with
  noisy labels. In: Proceedings of the IEEE/CVF Conference on Computer Vision
  and Pattern Recognition. pp. 7017--7025 (2019)

\bibitem{Weihao22MetaFormer}
Yu, W., Luo, M., Zhou, P., Si, C., Zhou, Y., Wang, X., Feng, J., Yan, S.:
  Metaformer is actually what you need for vision. In: Proceedings of the
  IEEE/CVF Conference on Computer Vision and Pattern Recognition (2022)

\bibitem{yuan2021multimodal}
Yuan, X., Lin, Z., Kuen, J., Zhang, J., Wang, Y., Maire, M., Kale, A., Faieta,
  B.: Multimodal contrastive training for visual representation learning. In:
  Proceedings of the IEEE/CVF Conference on Computer Vision and Pattern
  Recognition. pp. 6995--7004 (2021)

\bibitem{yuan2021activematch}
Yuan, X., Li, Z., Wang, G.: Activematch: End-to-end semi-supervised active
  representation learning. arXiv preprint arXiv:2110.02521  (2021)

\bibitem{zaheer2022generative}
Zaheer, M.Z., Mahmood, A., Khan, M.H., Segu, M., Yu, F., Lee, S.I.: Generative
  cooperative learning for unsupervised video anomaly detection. In:
  Proceedings of the IEEE/CVF Conference on Computer Vision and Pattern
  Recognition. pp. 14744--14754 (2022)

\bibitem{zaheer2020claws}
Zaheer, M.Z., Mahmood, A., Astrid, M., Lee, S.I.: Claws: Clustering assisted
  weakly supervised learning with normalcy suppression for anomalous event
  detection. In: European Conference on Computer Vision. pp. 358--376. Springer
  (2020)

\bibitem{zenati2018adversarially}
Zenati, H., Romain, M., Foo, C.S., Lecouat, B., Chandrasekhar, V.:
  Adversarially learned anomaly detection. In: 2018 IEEE International
  conference on data mining (ICDM). pp. 727--736. IEEE (2018)

\bibitem{zhan2019exploring}
Zhan, Y., Yu, J., Yu, T., Tao, D.: On exploring undetermined relationships for
  visual relationship detection. In: Proceedings of the IEEE/CVF Conference on
  Computer Vision and Pattern Recognition. pp. 5128--5137 (2019)

\bibitem{zhan2020multi}
Zhan, Y., Yu, J., Yu, T., Tao, D.: Multi-task compositional network for visual
  relationship detection. International Journal of Computer Vision
  \textbf{128}(8),  2146--2165 (2020)

\bibitem{zhan2018comprehensive}
Zhan, Y., Yu, J., Yu, Z., Zhang, R., Tao, D., Tian, Q.: Comprehensive
  distance-preserving autoencoders for cross-modal retrieval. In: Proceedings
  of the 26th ACM international conference on Multimedia. pp. 1137--1145 (2018)

\bibitem{zhang2022vsa}
Zhang, Q., Xu, Y., Zhang, J., Tao, D.: Vsa: Learning varied-size window
  attention in vision transformers. arXiv preprint arXiv:2204.08446  (2022)

\bibitem{zhao2017spatio}
Zhao, Y., Deng, B., Shen, C., Liu, Y., Lu, H., Hua, X.S.: Spatio-temporal
  autoencoder for video anomaly detection. In: Proceedings of the 25th ACM
  international conference on Multimedia. pp. 1933--1941 (2017)

\bibitem{zhong2021graph}
Zhong, H., Wu, J., Chen, C., Huang, J., Deng, M., Nie, L., Lin, Z., Hua, X.S.:
  Graph contrastive clustering. arXiv preprint arXiv:2104.01429  (2021)

\bibitem{zhou2017anomaly}
Zhou, C., Paffenroth, R.C.: Anomaly detection with robust deep autoencoders.
  In: Proceedings of the 23rd ACM SIGKDD international conference on knowledge
  discovery and data mining. pp. 665--674 (2017)

\end{thebibliography}
\end{document}